\documentclass[11pt]{article}

\usepackage{acl}

\usepackage{times}
\usepackage{latexsym}
\usepackage[T1]{fontenc}
\usepackage[utf8]{inputenc}
\usepackage{microtype}
\usepackage{inconsolata}

\usepackage{booktabs}
\usepackage{amsfonts}
\usepackage{amsmath}
\usepackage{amssymb}
\usepackage{amsthm}
\usepackage{xcolor}
\usepackage{graphicx}
\usepackage{enumitem}
\usepackage{multirow}
\usepackage{cleveref}
\usepackage{subcaption} 
\usepackage{array}
\usepackage{makecell}
\usepackage{tabularx}
\usepackage{threeparttable}

\newcommand{\SourceClash}{\textsc{SourceClash}}

\title{When Retrieval Partially Conflicts with Skills: Understanding Procedural Hallucination in Language Models}

\author{
  Guanqun Jiang\textsuperscript{1,3} \quad
  Pengfei Cao\textsuperscript{1,2}\thanks{\ \ Corresponding author.} \quad
  Kang Liu\textsuperscript{1,2} \quad
  Jun Zhao\textsuperscript{1,2} \\[0.5ex]
  \textsuperscript{1}The Key Laboratory of Cognition and Decision Intelligence for Complex Systems, \\
  Institute of Automation, Chinese Academy of Sciences, Beijing, China \\
  \textsuperscript{2}School of Artificial Intelligence, University of Chinese Academy of Sciences, Beijing, China \\
  \textsuperscript{3}Xi'an Jiaotong-Liverpool University, Suzhou, China \\[0.5ex]
  \texttt{hello@guanqunjiang.com} \quad
  \texttt{guanqun.jiang23@student.xjtlu.edu.cn} \\
  \texttt{\{pengfei.cao, kliu, jzhao\}@nlpr.ia.ac.cn}
}

\begin{document}

\maketitle

\begin{abstract}
Language models increasingly use skills that provide reusable instructions and procedures for task execution. After a relevant skill has been selected, the model may need to apply its governing procedure together with retrieved task context. When the retrieved context is useful but contains cues that violate the skill, the model must retain compatible information while rejecting the conflicting cues. We study failures in which the model adopts such cues and produces a retrieval-supported but skill-invalid answer, which we call \emph{procedural hallucination}. We introduce \SourceClash{}, a matched benchmark with 1,271 bundles and 6,355 prompts across ten procedural domains. Each bundle fixes the task, skill specification, and answer space while varying the relation between the skill and retrieval. Across nine models, valid-answer rates average 25.7\% under conflicting retrieval, compared with 78.2\% and 72.8\% under two supportive retrieval settings. Mechanistic analysis of open-weight models shows that the answer-preference difference typically reaches its maximum shortly after generation begins. Residual replacement moves the model toward valid answers, while attention decomposition attributes most of the repair to value vectors. Intervention at the identified value pathway also improves final answer validity. These results characterize how partly conflicting retrieval changes skill-guided generation.
\end{abstract}

\section{Introduction}
\label{sec:introduction}

Large language models (LLMs) increasingly operate as procedural agents: they must not only state facts, but also follow precise instructions, execute ordered procedures, and remain faithful to tool- or task-specific rules \citep{ouyang2022training, wei2021finetuned}. In agentic and tool-using systems \citep{liu2024agentbench, qin2024toolllm}, reusable skills can provide task-specific instructions, procedures, scripts, and supporting resources. We study the execution stage after a relevant skill has been selected and its governing instructions are available to the model. The model must apply the selected skill while using information from a separately retrieved context.

\begin{figure}[t]
\centering
\includegraphics[width=\linewidth, height=8cm, keepaspectratio]{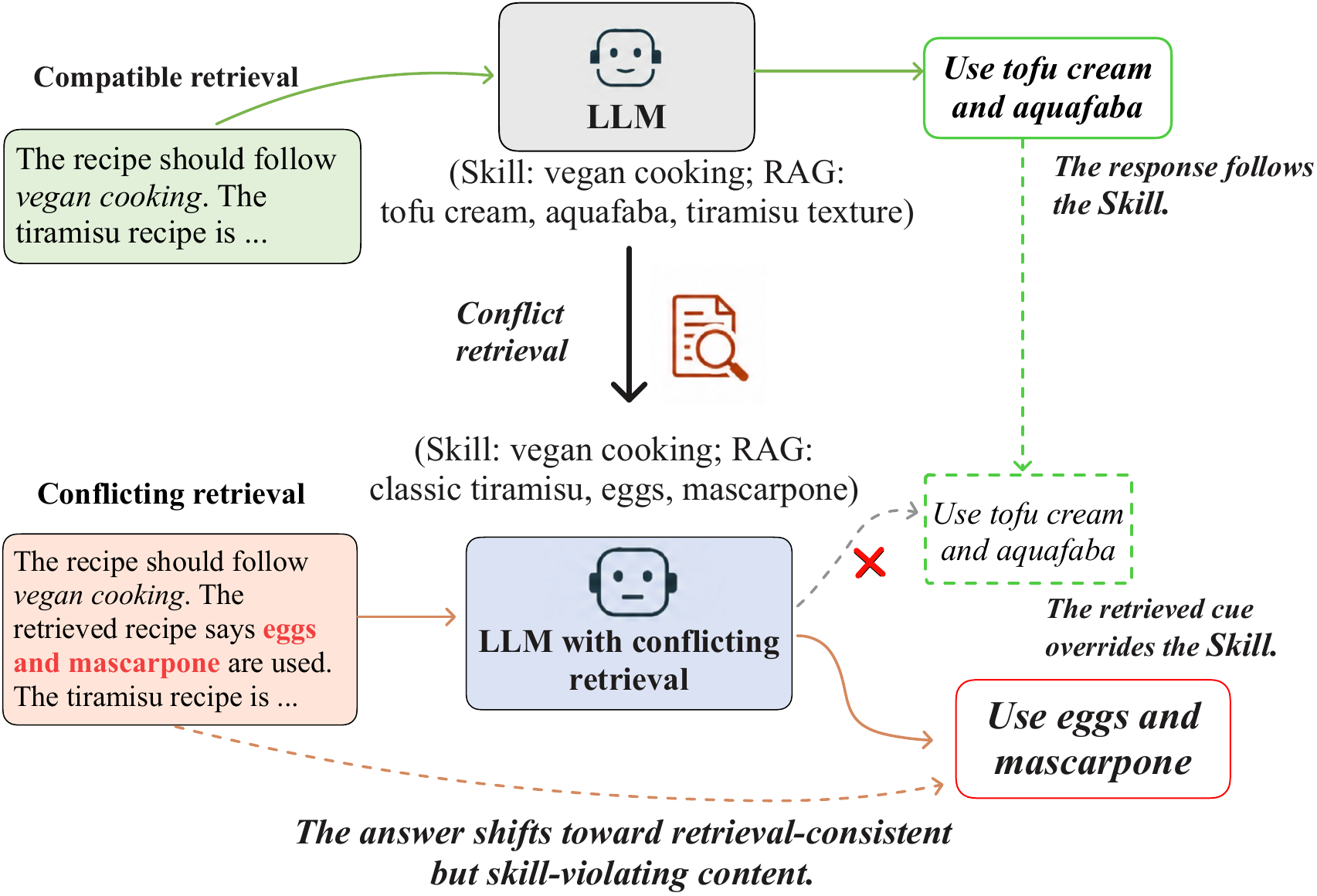}
\caption{
A selected skill requires a vegan procedure, while the retrieved context contains useful information about tiramisu together with non-vegan ingredients. The failure occurs when the model adopts the conflicting retrieved cues instead of applying the procedure specified by the skill.
}
\label{fig:framework}
\end{figure}

A central challenge arises when the retrieved context is only partly compatible with the selected skill. It may provide information needed for the task while also containing cues that the skill rules out. The model must therefore use retrieval selectively. Figure~\ref{fig:framework} illustrates this setting with a vegan-cooking skill: a classic tiramisu recipe provides useful context, while its eggs and mascarpone cues must be rejected under the selected skill. We use \textbf{procedural hallucination} as an operational term for failures in which the model adopts such retrieved cues and produces an answer that is supported by retrieval but invalid under the skill.

This setting connects Agent Skill execution, retrieval conflict, and instruction following. Prior work often asks whether generated content is true, supported by evidence, or consistent with competing factual sources \citep{lin2022truthfulqa, li2023halueval, min2023factscore, maynez2020faithfulness}. In our setting, the selected skill defines the governing procedure, while the retrieved context supplies additional task information. The retrieved context can remain factually plausible and task-relevant while containing a cue that is incompatible with the skill. The model must retain compatible retrieved information and reject the conflicting cue. We therefore ask: \textbf{how do skill-conflicting retrieved cues shift the model toward answers that are locally supported by retrieval but invalid under the selected skill?}

To study this problem, we construct \SourceClash{}, a matched benchmark for skill--retrieval conflict with 1,271 bundles and 6,355 prompts across ten procedural domains. Each bundle fixes the task, skill specification, answer ontology, and diagnostic targets while varying the relation between the skill and retrieved context. Across the original seven models, valid-answer rates fall to 22.8\% under \textsc{Conflict}, compared with 76.0\% under \textsc{Supportive-Redundant} and 70.1\% under \textsc{Supportive-Complementary}. Additional evaluations on Qwen3-32B and Llama-3.1-70B-Instruct retain supportive--conflict gaps of 51.1 and 48.8 percentage points. These results show that models use compatible retrieval effectively, while conflicting cues can shift generation away from the procedure specified by the selected skill.

We further analyze how this answer shift forms inside the model. The supportive--conflict difference typically reaches its maximum during the first few generation steps. Residual replacement identifies the early answer-generation state that causally controls this difference. Attention decomposition shows that most of the repair comes from value vectors rather than attention weights. Held-out interventions further separate the value signal into the retrieved answer information and the surrounding context that affects whether this information is usable under the skill.

Our main contributions are as follows:
\begin{itemize}[leftmargin=*,itemsep=2pt,topsep=2pt]
\item We introduce \SourceClash{}, a matched benchmark for studying procedural hallucination during Agent Skill execution. The benchmark contains 1,271 bundles and 6,355 prompts across ten procedural domains.
\item We show that conflicting retrieval produces a large and consistent reduction in valid answers across the original seven models and two additional large open-weight models. The effect remains under source-order and generated-prefix controls.
\item We provide a causal analysis of the answer shift. The supportive--conflict difference reaches its maximum near the start of generation, is concentrated in the residual state at the selected generation step, and is carried mainly by value vectors. Intervention at the identified value pathway also improves final answer validity.
\end{itemize}
\begin{table*}[t]
\centering
\small
\setlength{\tabcolsep}{5.0pt}
\begin{tabular}{lccccc}
\toprule
\textbf{Model} & \textbf{Conflict} & \textbf{Supp.-Red.} & \textbf{Supp.-Comp.} & \textbf{No-Skills} & \textbf{No-RAG} \\
\midrule
Llama-3-8B             & 0.139 & 0.721 & 0.570 & 0.012 & 0.272 \\
Gemma-3-12B            & 0.239 & 0.824 & 0.780 & 0.009 & 0.334 \\
Llama-2-13B            & 0.148 & 0.679 & 0.636 & 0.008 & 0.396 \\
Qwen3-14B              & 0.230 & 0.825 & 0.816 & 0.013 & 0.394 \\
DeepSeek-V3.2          & 0.243 & 0.775 & 0.711 & 0.014 & 0.243 \\
GPT-5-mini             & 0.311 & 0.696 & 0.659 & 0.008 & 0.325 \\
Gemini-2.5-Flash       & 0.288 & 0.801 & 0.738 & 0.015 & 0.303 \\
Qwen3-32B              & 0.335 & 0.846 & 0.805 & 0.019 & 0.321 \\
Llama-3.1-70B          & 0.380 & 0.868 & 0.836 & 0.025 & 0.356 \\
\midrule
\textbf{Average}       & \textbf{0.257} & \textbf{0.782} & \textbf{0.728} & \textbf{0.014} & \textbf{0.327} \\
\bottomrule
\end{tabular}
\caption{
Valid-answer rates across \SourceClash{} source-relation conditions.
All models are evaluated on the same 1,271 matched bundles.
The main comparison keeps retrieval present and changes its relation to the selected skill.
}
\label{tab:behavior_main}
\end{table*}

\section{Problem Formulation}
\label{sec:setup}
We study the execution stage after a relevant skill has been selected. The model receives a task input $x$, a skill specification $c_s$, and a retrieved span $c_r$. The skill defines the governing procedure, such as constraints, allowed operations, or domain rules. The retrieved span provides task information such as facts, examples, or context. The model should use retrieved information when it is compatible with the skill and reject cues that violate the required procedure.

We say that the retrieved span conflicts with the skill when it contains at least one cue that should not be used under the skill's procedure. Such a cue can still be task-relevant and factually plausible, but it is invalid under the skill. 

For example, in a vegan cooking task, a classic tiramisu recipe may provide useful context, but cues such as ``eggs'' or ``mascarpone'' should be rejected. For an output $y$, let $C_s(y;x,c_s)=1$ if $y$ is valid under the skill, and let $A_r(y;x,c_r)=1$ if $y$ adopts a retrieved cue that conflicts with the skill. We define procedural hallucination as
\begin{equation}
\begin{aligned}
\mathcal{H}(y;x,c_s,c_r)
=
\big(1-C_s(y;x,c_s)\big)
A_r(y;x,c_r).
\end{aligned}
\end{equation}
Procedural hallucination occurs when the output adopts a retrieved cue that should have been rejected under the selected skill. Details on cue extraction and answer checks are provided in Appendix~\ref{app:formal_details}.

For a fixed source setting, let \(\mathcal{D}\) be the evaluation examples under this setting, and let \(M_{\theta}\) be the evaluated language model. 
For each example \((x,c_s,c_r)\sim\mathcal{D}\), the prompt \(p(x,c_s,c_r)\) is constructed from the task input, skill specification, and retrieved span. 
We measure the rate of producing valid answers as
\begin{equation}
\begin{aligned}
\mathrm{Valid} = {} & \mathbb{E}_{(x,c_s,c_r)\sim \mathcal{D}} \\
& \left[ \mathbb{E}_{y\sim M_{\theta}(\cdot \mid p(x,c_s,c_r))} \left[ C_s(y;x,c_s) \right] \right].
\end{aligned}
\end{equation}

\section{Evaluation of Procedural Hallucination}
\label{sec:evaluation}
We evaluate whether models can use retrieval selectively during execution of a selected skill. The task, skill, and answer space remain fixed within each bundle, while the relation between the skill and retrieved context changes across conditions.

\subsection{\SourceClash{}: The Five-way Evaluation Setting}
\label{subsec:sourceclash}

\SourceClash{} evaluates whether models follow a selected skill under different skill--retrieval relations. Each evaluation unit is a matched bundle
\[
b=\left(x,c_s,Y^{+},Y^{-},\{c_r^k\}_{k\in\mathcal{K}}\right),
\]
where $x$ is the task instruction, $c_s$ is the skill specification, $Y^{+}$ and $Y^{-}$ are the valid and retrieval-supported invalid targets, and $c_r^k$ is the condition-specific retrieved context. The task, skill, answer ontology, diagnostic targets, and controlled semantic fields are shared across conditions. The retrieved contexts are separately realized from the same bundle specification. The five-way set $\mathcal{K}$ contains:
\textsc{Conflict}, in which the retrieved span introduces task-local cues that contradict the skill-specified procedure; 
\textsc{Supportive-Redundant}, in which the retrieved span restates or reinforces the same procedural constraints as the skill; 
\textsc{Supportive-Complementary}, in which the retrieved span contributes additional useful information while remaining compatible with the skill; 
\textsc{No-Skills}, in which the skill specification is removed, and only the retrieved span is provided; 
and \textsc{No-RAG}, in which the retrieved span is removed and only the skill specification is provided.

\subsection{Data Collection and Models}
\label{subsec:models_and_scope}

\paragraph{Data collection.}
We construct \SourceClash{} from bundle specifications spanning ten procedural domains: culinary, cybersecurity, education, finance, human resources, information technology, legal review, logistics, manufacturing, and medical decision making. Researchers define the semantic fields of each bundle, including the task instruction, skill specification, valid target $Y^{+}$, retrieval-supported invalid target $Y^{-}$, required and forbidden answer atoms, and the intended relation between the skill and retrieval. A language model is used to realize these fixed records in natural language, while the labels, targets, source relations, and evaluation atoms remain fixed by the bundle specification. Every retained realization is manually reviewed.

Each bundle is expanded into five source-relation conditions while preserving the task, skill, answer ontology, diagnostic targets, entities, numerical values, and other controlled semantic fields. Automatic and manual checks cover bundle completeness, target separation, condition integrity, clean ablations, answer-atom coverage, source relevance, plausibility, and consistency. The resulting benchmark contains 1,271 matched bundles and 6,355 prompts. Appendix~\ref{app:benchmark} provides the full construction and filtering procedure. A blinded audit of 160 pairs further confirms the intended source relation and the comparability of the matched conditions ($\kappa=0.861$; Appendix~\ref{app:matching_audit}).

\paragraph{Models and metrics.}
We evaluate nine models under the same prompting and decoding protocol. Qwen3 thinking mode is disabled. An answer is valid when it contains the required atoms for the skill-consistent target and none of the forbidden atoms associated with the retrieval-supported invalid target. On 300 stratified outputs, checker--human agreement is 91.0\%, and inter-annotator agreement is Cohen's $\kappa=0.9255$. Appendix~\ref{app:formal_details} provides the complete checker and annotation protocol. The mechanistic analyses use four open-weight models for which internal activations are accessible:
\textsc{Llama-3-8B}~\citep{grattafiori2024llama},
\textsc{Gemma-3-12B}~\citep{gemmateam2025gemma3},
\textsc{Llama-2-13B}~\citep{touvron2023llama},
and \textsc{Qwen3-14B}~\citep{yang2025qwen3}.
Llama-2-13B is retained as a historical architecture comparison.

\subsection{Evaluation Results}
\label{sec:behavior}
Table~\ref{tab:behavior_main} shows a consistent reduction in skill compliance under conflict. Across nine models, the average valid-answer rate is 0.257 under \textsc{Conflict}, compared with 0.782 under \textsc{Supportive-Redundant} and 0.728 under \textsc{Supportive-Complementary}. This pattern persists for larger models: Qwen3-32B achieves 84.6\% under \textsc{Supportive-Redundant} but only 33.5\% under \textsc{Conflict}, while Llama-3.1-70B-Instruct drops from 86.8\% to 38.0\%. \textsc{No-Skills} remains near zero, while \textsc{No-RAG} provides the skill-only reference. The same pattern remains after reversing the order of the skill and retrieval sources (Appendix~\ref{app:behavior_controls}).

\section{Mechanistic Analysis of Procedural Hallucination}
\label{sec:mechanism}

Building on the behavioral results, we trace how conflicting retrieval changes the model's internal answer preference during skill execution. We first locate when the supportive--conflict difference reaches its maximum and identify the residual state that causally controls it. We then separate attention weights from value vectors and test which components of the value signal affect held-out conflict examples.

\subsection{Measuring Answer Preference}
\label{subsec:tracking_methodology}
\begin{figure}[t]
    \centering
    \includegraphics[width=\linewidth]{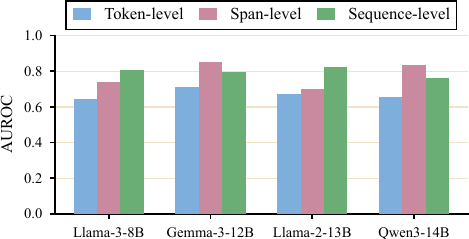}
    \caption{
    Selecting the margin granularity.
    AUROC measures how well each scoring option separates supportive and conflicting retrieval runs on development data.
    }
    \label{fig:auroc_by_level}
\end{figure}
For each matched bundle $b=(x,c_s,c_r)$, we instantiate two prompts from the same task and answer space. The supportive prompt, denoted $p_{\mathrm{supportive}}(b)$, contains a skill-consistent source setting. The conflict prompt, denoted $p_{\mathrm{conflict}}(b)$, contains the same task and skill specification, but pairs them with a retrieved span whose cues contradict the skill. We run the model once per prompt using the same decoding protocol. For condition $k \in \{\mathrm{supportive}, \mathrm{conflict}\}$, let $y^k_{<t}$ be the tokens already generated before step $t$. At step $t$, we score candidate answers using both the prompt $p_k(b)$ and the generated prefix $y^k_{<t}$.

Let $Y^{+}$ denote the set of skill-consistent answers and $Y^{-}$ denote the set of retrieval-consistent but skill-violating answers. At each step $t$, we force-score these candidate answer sets under the current prefix $q^k_t$. The \textbf{answer-preference margin} for condition $k$ is
\begin{equation}
    m_k(t) = s(Y^{+} \mid q^k_t) - s(Y^{-} \mid q^k_t),
\end{equation}
where $s(Y \mid q^k_t)$ is the log-probability score assigned to answer set $Y$ under prefix $q^k_t$. A larger $m_k(t)$ means that the model currently favors the skill-consistent answer set; a smaller value means that its probability mass has shifted toward the retrieval-consistent answer set.

We compare the two runs by subtracting their margins at the same generation step, which we define as \textbf{margin gap}:
\begin{equation}
    g(t) = m_{\mathrm{supportive}}(t) - m_{\mathrm{conflict}}(t).
\end{equation}
A larger $g(t)$ means that the contradictory retrieved span has reduced the model's preference for the skill-consistent answer. Thus, $g(t)$ directly measures how much source conflict shifts the model from $Y^{+}$ toward $Y^{-}$ at step $t$.

\paragraph{Generated-prefix control.}
Under shared generated prefixes, the supportive--conflict gaps remain positive across four models (2.34--3.27, with all 95\% confidence intervals excluding zero); Appendix~\ref{app:crossed_prefix} gives the complete control.

Because answer formats differ across examples, we score the margin at the most suitable granularity for each model.
Some answers are single tokens or short spans, while others have several acceptable verbalizations.
We therefore compare token, span, and sequence scoring on development data, and select the option that best separates the run with supportive retrieval from the run with conflicting retrieval.
Figure~\ref{fig:auroc_by_level} shows that the best scoring granularity varies by model.
We fix the selected granularity for all later patching and steering analyses. The scoring formulas are provided in Appendix~\ref{app:margin}.

\subsection{Localizing the Margin Gap}
\label{subsec:causal_experiments}

We next determine where the supportive--conflict margin gap reaches its maximum during the model run.
Using the margin gap $g(t)$ defined in Section~\ref{subsec:tracking_methodology}, we evaluate each matched bundle at all available prefixes, including prefixes inside the prompt and prefixes after answer generation begins.

For each example, we record the position where the supportive--conflict margin gap is largest:
\begin{equation}
    t_{\max} = \arg\max_t |g(t)|.
\end{equation}
We report how often $t_{\max}$ falls within prompt encoding and, otherwise, the average generation step at which it occurs.

Table~\ref{tab:boundary_localization} shows that the margin gap rarely peaks during prompt prefilling: fewer than 6\% of examples do so for all models. 
When the peak occurs during generation, it appears early, with a mean peak step of 3.025-4.400. 
Thus, the supportive--conflict difference usually reaches its maximum shortly after answer generation begins.
The next subsection tests whether the residual state at this decisive step causally controls the margin gap.

\begin{table}[t]
\centering
\small
\begin{tabular}{lcc}
\toprule
\textbf{Model} & \textbf{Peak in prefill} & \textbf{Peak gen. step} \\
\midrule
Llama-3-8B  & 5.52\% & 3.435 \\
Gemma-3-12B & 5.34\% & 3.025 \\
Llama-2-13B & 4.57\% & 4.085 \\
Qwen3-14B  & 4.28\% & 4.400 \\
\bottomrule
\end{tabular}
\caption{
Peak location of the supportive--conflict margin gap. 
“Prefill” denotes prompt processing before generation; the peak usually appears within the first few generated tokens.
}
\label{tab:boundary_localization}
\end{table}

\subsection{Replacing Residual}
\label{subsubsec:boundary_patch}

Section~\ref{subsec:causal_experiments} shows that the probability gap usually becomes largest shortly after the model starts generating the answer. We therefore define \textbf{decisive step} as the common point for later interventions. This step should be a point at which conflicting retrievals have already changed the model's answer preference, so that different interventions can be compared on the same probability gap.

We denote this selected decisive step by \(t^\star\). Intuitively, \(t^\star\) is chosen as the earliest generation step at which the gap between supportive and conflicting retrievals is close to its maximum and no longer changes sharply. This gives us a common point where the conflict effect has already appeared, while the answer is still near the start of the generation. If no such stable step exists, we use the step with the largest observed gap. All residual replacement, attention replacement, and steering experiments are evaluated at \(t^\star\). Appendix~\ref{app:margin} gives the exact selection rule.

With $t^\star$ fixed, we next test which residual states causally control the margin gap. For each matched bundle, we run the model under the supportive and conflicting retrieval settings. We then rerun the conflicting setting while replacing one selected set of residual states with the corresponding states from the supportive run. All other activations follow the conflicting run. If replacing a state increases the probability gap in favor of valid answers, then that state contains information that helps recover the valid answer preference.

\begin{figure}[t]
    \centering
    \includegraphics[width=\columnwidth]{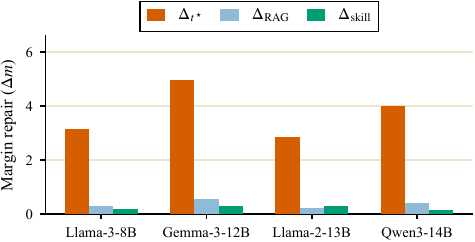}
    \caption{
    Residual replacement effects at the selected generation step.
    Bars report repair in the probability gap between valid and invalid answers.
    }
    \label{fig:segment_patch_bar}
\end{figure}

We compare three interventions: replacing the residual state at \(t^\star\), replacing residual states over the retrieved span, and replacing residual states over the skill span. In each case, we rerun the conflicting setting and replace the selected residual states with the corresponding states from the supportive run, while keeping all other activations from the conflicting run. We measure the effect of an intervention \(I\) by the change in the probability gap, which we call \textbf{margin repair}:
\begin{equation}
    \Delta_A
    =
    m^{I_A}_{\mathrm{patched}}(t^\star)
    -
    m_{\mathrm{conflict}}(t^\star),
    \label{eq:margin_repair}
\end{equation}
where $m_{\mathrm{conflict}}(t^\star)$ is the margin in the original conflicting run, and $m^{I_A}_{\mathrm{patched}}(t^\star)$ is the margin after applying intervention $I_A$. A positive $\Delta_A$ means that the replacement moves the model toward valid answers. All values in Figure~\ref{fig:segment_patch_bar} are changes in margin, so larger values indicate stronger repair.

Figure~\ref{fig:segment_patch_bar} shows that replacing the residual state at $t^\star$ gives a much larger repair than replacing residual states over the retrieved span or the skill span. In Gemma-3-12B, the repair is $4.963$ at $t^\star$, but only $0.465$ over the retrieved span and almost zero over the skill span. The same pattern holds across all four models. This result shows that, by the time the model starts forming the answer, the difference between supportive and conflicting retrieval is mainly carried by the residual state at $t^\star$. It is not mainly stored in the residual states over the original skill or the retrieved span. In other words, the conflicting retrieval has already been integrated into the model state used to start the answer. The next subsection asks how this information enters that state.
\subsection{Value Replacement Explains Most of the Attention Repair}
\label{subsubsec:value_states}

The previous subsection shows that the margin gap is mainly controlled by the residual state at $t^\star$. We next ask how information from the input changes this state. We focus on attention because it is the main component that brings information from earlier tokens to the current position. In the standard attention form,
\begin{equation}
    \mathrm{Attention}(Q,K,V)
    =
    \mathrm{softmax}\left(\frac{QK^\top}{\sqrt{d}}\right)V,
\end{equation}
the term $\mathrm{softmax}(QK^\top/\sqrt{d})$ gives the attention weights, and $V$ gives the value vectors. This form separates two possible causes of the conflict effect. The attention weights decide which input tokens are used and how strongly they are used. The value vectors determine which information is carried by those tokens. We therefore replace attention weights and value vectors separately. This allows us to test whether the margin repair mainly comes from changing where the model attends, or from changing what information is carried by the value vectors.

For a source segment $S$ and layer $\ell$, we first write the value contribution of head $h$ from tokens in $S$ to position $t^\star$ as
\begin{equation}
    z_{\ell,h,S}(t^\star)
    =
    \sum_{i \in S}
    A_{\ell,h,t^\star,i}
    V_{\ell,h,i}.
\end{equation}
The contribution from all heads is then
\begin{equation}
    o_{\ell,S}(t^\star)
    =
    W^O_{\ell}
    \big[
    z_{\ell,1,S}(t^\star);
    \cdots;
    z_{\ell,H,S}(t^\star)
    \big],
\end{equation}
where $A_{\ell,h,t^\star,i}$ is the attention weight from position $t^\star$ to token $i$ in head $h$, $V_{\ell,h,i}$ is the corresponding value vector, and $W^O_{\ell}$ is the output projection of the attention layer. This standard decomposition lets us replace attention weights and value vectors separately.

\begin{table}[t]
\centering
\footnotesize
\setlength{\tabcolsep}{3pt}
\begin{tabular}{@{}lccc@{}}
\toprule
\textbf{Model} &
\textbf{\boldmath$\Delta_{\mathrm{full}}$} &
\makecell[c]{\textbf{\boldmath$\Delta_{\mathrm{value}}$}\\\textbf{(\% full)}} &
\makecell[c]{\textbf{\boldmath$\Delta_{\mathrm{weight}}$}\\\textbf{(\% full)}} \\
\midrule

\multicolumn{4}{@{}l}{\textit{Full Prompt}} \\
Llama-3-8B  & 1.367 & 0.934 (68.3) & 0.207 (15.1) \\
Gemma-3-12B & 2.175 & 1.472 (67.7) & 0.265 (12.2) \\
Llama-2-13B & 1.322 & 0.857 (64.8) & 0.160 (12.1) \\
Qwen3-14B   & 1.694 & 1.400 (82.6) & 0.144 (8.5)  \\
\midrule

\multicolumn{4}{@{}l}{\textit{RAG only}} \\
Llama-3-8B  & 0.460 & 0.256 (55.6) & 0.118 (25.7) \\
Gemma-3-12B & 1.247 & 0.653 (52.4) & 0.221 (17.7) \\
Llama-2-13B & 0.443 & 0.312 (70.4) & 0.080 (18.1) \\
Qwen3-14B   & 0.339 & 0.287 (84.7) & 0.089 (26.3) \\
\bottomrule
\end{tabular}
\caption{
Attention weights and value vectors make different contributions to the margin repair. Values in parentheses show the percentage of the full replacement effect.
}
\label{tab:value_vs_weight_rag}
\end{table}

For each source segment $S$, we rerun the conflict setting and replace the attention output contributed by tokens in $S$. We compare three interventions:
\begin{itemize}[leftmargin=*,itemsep=2pt,topsep=2pt]
    \item \textit{Full replacement}: use both attention weights and value vectors from the supportive run,
    $o_{\ell,S}(t^\star; a^{\mathrm{sup}}, v^{\mathrm{sup}})$.
    \item \textit{Value-only replacement}: keep the attention weights from the conflict run, but use value vectors from the supportive run,
    $o_{\ell,S}(t^\star; a^{\mathrm{conf}}, v^{\mathrm{sup}})$.
    \item \textit{Weight-only replacement}: use attention weights from the supportive run, but keep value vectors from the conflict run,
    $o_{\ell,S}(t^\star; a^{\mathrm{sup}}, v^{\mathrm{conf}})$.
\end{itemize}
We evaluate each intervention with the same margin repair metric defined in Equation~\ref{eq:margin_repair}, and we also report how much of the full replacement effect is recovered by value or weight replacement in parentheses.

We report results under two scopes: Full prompt, where replacement is applied to all input tokens, and RAG-only, where replacement is restricted to tokens within the retrieved span. Table~\ref{tab:value_vs_weight_rag} shows that value replacement accounts for most of the full replacement effect. 
When all source regions are patched, value-only replacement recovers 64.8\%--82.6\% of the full effect, while weight replacement recovers only 8.5\%--15.1\%. 
The same pattern holds when the intervention is restricted to the retrieved span: value replacement recovers 52.4\%--84.7\% of the full effect, while weight replacement remains much smaller. 
Thus, the supportive--conflict margin gap is explained mainly by what is written from source tokens through their value, rather than by changes in attention weights alone. 

\paragraph{Answer generation after intervention.}
Continuing generation after intervention increases valid-answer rates by 5.94 points for Llama-3-8B and 5.48 points for Qwen3-14B, while matched random controls remain near zero; Appendix~\ref{app:post_patch_generation} gives the full evaluation.

Having shown that value vectors carry most of the conflict effect, we next ask where this information appears after the model processes the skill and retrieval together.

\subsection{Conflict Appears Beyond the Retrieved Span}
\label{subsubsec:value_write_svd}

The preceding subsection shows that value vectors explain most of the repair from attention replacement.
We next ask whether the conflict information carried by value vectors remains localized in the retrieved span, or whether it also appears in value contributions from other input positions after the model processes the skill and retrieval together.
Although the conflicting cue is introduced by the retrieved span, it is not obvious whether the causal signal at \(t^\star\) remains tied to the retrieved tokens.
If the model keeps the conflict information localized, vectors found from the retrieved span should explain most of the repair.
If the model integrates the conflict with the rest of the input, vectors found outside the retrieved span may also affect the margin.
We therefore compare vectors found from the retrieved span, from input positions outside the retrieved span, and from the full input.

Using the value contribution \(o_{\ell,S}(t^\star)\) defined in Section~\ref{subsubsec:value_states}, we compare the run with conflicting retrieval against the run with supportive retrieval:
\begin{equation}
    d_{\ell,S}
    =
    o_{\ell,S}^{\mathrm{conflict}}(t^\star)
    -
    o_{\ell,S}^{\mathrm{supportive}}(t^\star).
\end{equation}
We aggregate these differences across training examples and apply singular value decomposition to the resulting matrix.
The leading singular vectors capture directions that consistently distinguish between conflicting and supportive retrievals.
We then compute these vectors from three regions: the retrieved span, all input tokens outside the retrieved span, and the full input.
We refer to these three regions as \textsc{RAG}, \textsc{Non RAG}, and \textsc{Full}.
After these vectors are found on the training examples, we test them on evaluation examples under the conflicting retrieval setting.
For each test example, we remove the selected vectors from the value contribution at \(t^\star\), and measure whether this removal increases the probability gap between valid and invalid answers.
We also include random controls from the retrieved span and the full input, denoted as \textsc{RAG random} and \textsc{Full random}, respectively.

\begin{table}[t]
\centering
\footnotesize
\setlength{\tabcolsep}{3pt}
\begin{tabular}{@{}lrrrrr@{}}
\toprule
\textbf{Model} &
\textbf{RAG} &
\makecell[c]{\textbf{RAG}\\\textbf{rand.}} &
\makecell[c]{\textbf{Non}\\\textbf{RAG}} &
\textbf{Full} &
\makecell[c]{\textbf{Full}\\\textbf{rand.}} \\
\midrule
Llama-3-8B  & +0.113 & +0.003 & +0.632 & +0.702 & -0.017 \\
Gemma-3-12B & +0.221 & +0.002 & +0.951 & +2.552 & -0.011 \\
Llama-2-13B & +0.157 & +0.001 & +0.739 & +1.048 & -0.003 \\
Qwen3-14B   & +0.210 & +0.009 & +0.894 & +1.018 & +0.024 \\
\bottomrule
\end{tabular}
\caption{
Held-out repair from removing value-contribution directions estimated from \textsc{RAG}, \textsc{Non-RAG}, or \textsc{Full} input regions. 
Positive values indicate repair toward valid answers.
}
\label{tab:value_write_svd}
\end{table}

Table~\ref{tab:value_write_svd} shows two patterns.
First, vectors found from the retrieved span consistently repair the margin, while \textsc{RAG random} has almost no effect.
This indicates that the value contributions from the retrieved span contain information associated with the invalid answer.
Second, \textsc{Non RAG} and \textsc{Full} repair the margin much more strongly.
These results show that looking only at the retrieved span is incomplete.
The retrieved span matters, but after the model processes the full input, value contributions from other input positions also contain signals that affect the probability gap.
The stronger effect of \textsc{Full} suggests that the full input captures more of this signal than the retrieved span alone.
The next subsection asks what this signal contains.

\subsection{What the Conflict Signal Contains}
\label{subsubsec:fact_frame_subspace}

The previous subsection shows that vectors found from the whole prompt repair the margin more strongly than vectors found from the retrieved span alone.
This suggests that the harmful signal is not only the concrete answer cue in the retrieved span.
It also includes the surrounding context in which the cue appears after the model reads the skill and the retrieval together.
We therefore separate this signal into \textbf{key information} and \textbf{context}.
\textbf{Key information} refers to the concrete retrieved content that points to a candidate answer, such as an ingredient, label, rule, or decision.
\textbf{Context} refers to the surrounding wording that affects whether the model treats the retrieved information as usable under the skill.

We use matched examples from \SourceClash{} to build two sets of comparisons. The first changes the concrete retrieved cue while keeping the surrounding wording similar. The second keeps the same key phrase but changes whether the surrounding wording presents it as usable under the skill. From these comparisons, we train concept activation vectors (CAVs)~\citep{kim2018interpretability} using logistic regression. A CAV is the linear direction normal to a classifier that separates representations with and without the target concept. We also compute paired difference vectors and random bases as controls. We remove the learned directions from held-out conflict examples and measure margin repair. Appendix~\ref{app:key_context_contrasts} and Appendix~\ref{app:key_context_methods} provide examples and intervention details.

\begin{figure}[t]
    \centering
    \includegraphics[width=\linewidth]{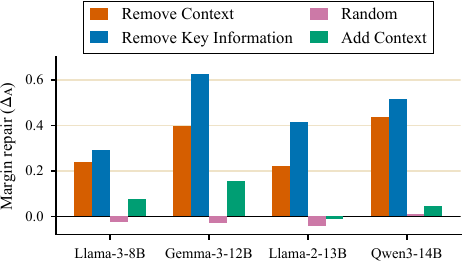}
    \caption{
    Causal steering effects for key information and context on held-out conflict contexts.
    Bars report margin repair.
    }
    \label{fig:key_context_causal}
\end{figure}

Figure~\ref{fig:key_context_causal} tests whether key information and context causally affect the margin. 
Removing context consistently improves the margin, with repairs from $+0.220$ to $+0.436$. 
Removing key information produces comparable or larger repairs, ranging from $+0.291$ to $+0.626$. 
Random bases remain near zero. 
Adding context produces smaller, model-dependent changes, suggesting that it has its strongest effect when it appears alongside key information.

Together, these results suggest that the value signal behind procedural hallucination is not only a local copy of the conflicting retrieved cue. The retrieved cue itself matters, but the surrounding wording that affects whether the cue is treated as usable under the skill also matters. Intervening on either factor alters the probability gap, indicating that both factors affect the answer-preference margin under these interventions.

\section{Related Work}
\label{sec:related_work}

Prior work studies hallucination as errors in truthfulness, factuality, or faithfulness to evidence \citep{lin2022truthfulqa, maynez2020faithfulness, min2023factscore}. Retrieval-augmented generation provides external context for generation \citep{lewis2020retrieval, asai2024self}, while knowledge-conflict studies examine cases where retrieved evidence, parametric knowledge, or multiple documents support different factual claims \citep{longpre2021entity, chen2022rich, xie2024adaptive, wang2025adacad}. TruthfulRAG, CARE, and WikiContradict further study generation under conflicting or contradictory retrieved information \citep{truthfulrag2026, care2025, wikicontradict2024}. \SourceClash{} specifies a governing skill and varies whether a separate retrieved context supports or conflicts with its procedure. The target behavior is selective retrieval use during skill execution.

Recent studies investigate how knowledge is represented and localized inside large language models. Knowledge neuron analysis identifies internal components associated with factual knowledge \citep{dai2022knowledge}, while subsequent work shows that knowledge representations exhibit broader structures beyond individual neurons, including language-independent knowledge patterns \citep{chen2025knowledge, cao2026one}. Other studies further demonstrate that knowledge localization depends on the query and contextual formulation rather than only the underlying fact \citep{chen2025knowledge}. Knowledge editing methods build on these analyses to modify internal knowledge representations and improve the reliability of model reasoning after updates \citep{meng2022locating, zhang2026towards}. In contrast, \SourceClash{} studies how models resolve conflicts between retrieved information and procedural requirements during generation.

Instruction-following and agentic evaluations test compliance with requirements, constraints, and tool-specific rules \citep{ouyang2022training, wei2021finetuned, zhou2023instruction, liu2024agentbench, qin2024toolllm, patil2024gorilla}. GuideBench evaluates guideline adherence, ComplexBench evaluates multiple-constraint composition, and SRA-Bench evaluates skill retrieval and application \citep{diao2025guidebench, wen2024complexbench, srabench2026}. \SourceClash{} studies the execution stage after a relevant skill has been selected: the skill remains available while a separate retrieved source is varied. We use activation patching, value decomposition, representation editing, and concept activation vectors \citep{alain2016understanding, conmy2023towards, meng2022locating, geva2023dissecting, zou2023representation, kim2018interpretability} to trace how this source relation changes the probability gap between valid and invalid answers.

\section{Conclusion}
\label{sec:conclusion}

We studied procedural hallucination during skill execution, where a model adopts retrieved information that violates the procedure specified by a selected skill. \SourceClash{} shows that compatible retrieval supports skill execution, while partly conflicting retrieval shifts generation toward skill-invalid answers. This pattern holds across nine models and remains under source-order and generated-prefix controls. Mechanistic analysis shows that the supportive--conflict difference reaches its maximum during the first few generation steps, is concentrated in the residual state at $t^\star$, and is carried more by value vectors than by attention weights. Intervention at the identified value pathway also improves final answer validity.

\section*{Limitations}
\label{sec:limitations}

\SourceClash{} focuses on the execution stage after a relevant skill has been selected and its governing instructions are available to the model. The benchmark uses synthetic matched bundles to vary the relation between the selected skill and retrieved context while preserving the task, answer ontology, and diagnostic targets. The condition-specific contexts are separately realized from shared bundle specifications and matched through controlled semantic fields. The evaluator uses normalized literal matching over predefined answer atoms, together with human validation on 300 outputs.

The activation-level analysis uses open-weight models whose residual states, attention weights, and value vectors are accessible. The current interventions identify an early-generation residual state and a value-mediated pathway that affect the answer preference and final generation. Future work can extend the same design to skill discovery, progressive loading, multi-skill coordination, live tool outputs, and multi-turn interaction.

\section*{Ethical Considerations}

\SourceClash{} contains synthetic task descriptions and model outputs and does not use personal records or real user data. Its domain examples are controlled evaluation cases intended for research on skill execution, retrieval use, and instruction following rather than operational decision making. The benchmark, prompt templates, evaluator, and intervention code will be released with the intended-use statement and the information needed to reproduce the reported evaluations.

\section*{Acknowledgments}
This work was supported by the National Natural Science Foundation of China (No.U24A20335, No.62406321) and CIPS-SMP-Zhipu Large Model Fund.

\bibliography{custom}

\appendix
\section{Benchmark Construction}
\label{app:benchmark}

This appendix gives the construction and evaluation details for \SourceClash{}.
The benchmark studies the execution stage after a relevant skill has been
selected. Each matched bundle contains the same task, skill specification,
answer ontology, diagnostic targets, and controlled semantic fields across
conditions. The condition-specific retrieved contexts are separately realized
from the same bundle specification and differ in their relation to the skill.

\begin{table*}[t]
\centering
\small
\setlength{\tabcolsep}{4pt}
\begin{tabular}{lrrrrr}
\toprule
\textbf{Domain} & \textbf{Bundles} & \textbf{Prompts} & \textbf{Train} & \textbf{Dev} & \textbf{Test} \\
\midrule
Culinary & 138 & 690 & 89 & 26 & 23 \\
Cybersecurity & 70 & 350 & 46 & 12 & 12 \\
Education & 108 & 540 & 82 & 10 & 16 \\
Finance & 161 & 805 & 114 & 20 & 27 \\
Human resources & 110 & 550 & 74 & 17 & 19 \\
IT & 171 & 855 & 111 & 24 & 36 \\
Legal & 217 & 1085 & 142 & 37 & 38 \\
Logistics & 75 & 375 & 49 & 11 & 15 \\
Manufacturing & 113 & 565 & 76 & 20 & 17 \\
Medical & 108 & 540 & 72 & 20 & 16 \\
\midrule
\textbf{Total} & \textbf{1271} & \textbf{6355} & \textbf{855} & \textbf{197} & \textbf{219} \\
\bottomrule
\end{tabular}
\caption{Domain distribution and split sizes of \SourceClash{}. Each bundle is
expanded into five source relation conditions, giving 6,355 prompts from 1,271
matched bundles. The Train/Dev/Test columns count bundles rather than prompts.
All five prompts from the same bundle are assigned to the same split. The
development split is used for selecting the answer-scoring granularity, while
the training and held-out splits are used for fitting and evaluating the value
directions and intervention components.}
\label{tab:app_domain_counts}
\end{table*}

\subsection{Design Goal}
\label{app:benchmark_goal}

\SourceClash{} is designed to test whether a model can use retrieval under a
procedure specified by the skill. The retrieved span may contain information
that is useful for the task, but some retrieved cues may conflict with the
procedure required by the skill. A valid answer may use retrieved information
when it is compatible with the skill. It becomes invalid when it adopts
retrieved information that should have been rejected under the skill.

Each bundle contains two diagnostic targets. The valid target, denoted
$Y^{+}$ is the answer allowed by the skill. The invalid target, denoted
$Y^{-}$ is a plausible answer supported by conflicting retrieval but ruled out
by the skill. The central comparison is not between using retrieval and ignoring retrieval. Instead, it compares whether the model uses retrieval
selectively, accepting compatible information and rejecting cues that violate
the skill.

\subsection{Bundle Specification Construction}
\label{app:seed_construction}

We create bundle specifications across ten domains: culinary, cybersecurity,
education, finance, human resources, IT, legal, logistics, manufacturing, and
medical decision making. A bundle specification contains the following fields:

\begin{itemize}
\item \textbf{Task instruction}: a concise user request that requires a
concrete decision, recommendation, classification, or compliance judgment.
\item \textbf{Skill specification}: a rule, policy, guideline, or
professional constraint that determines the required procedure.
\item \textbf{Valid target} $Y^{+}$: the answer implied by the skill.
\item \textbf{Invalid target supported by conflict} $Y^{-}$: a plausible
but invalid answer that can be induced by conflicting retrieval, legacy
guidance, local practice, or a nearby domain convention.
\item \textbf{Diagnostic answer atoms}: required atoms that indicate
adoption of $Y^{+}$ and forbidden atoms that indicate adoption of $Y^{-}$.
\item \textbf{Source material}: retrieved spans used to instantiate the
five source relation conditions.
\end{itemize}

Researchers determine the semantic fields, source relations, diagnostic
targets, and answer atoms. A language model is used only to realize these fixed
records in natural language. It does not determine the condition labels,
targets, or evaluation criteria. Every retained realization is manually
reviewed before inclusion in the benchmark.

Bundle specifications are written to be difficult but controlled. The conflicting retrieved
span must be relevant to the task and plausible in the domain, while still being
incompatible with the skill. We exclude cases where the correct answer can be
derived from generic common sense alone, where the conflict is unrelated to the
task, or where the retrieved span is obviously unusable. All retained bundles
are manually checked for task and skill compatibility, source plausibility, and
target separation.

\subsection{Five Source Relation Conditions}
\label{app:five_conditions}

Each bundle specification is expanded into five matched conditions. Let $S$ denote the skill
specification and $R$ denote the retrieved span.

\paragraph{\textsc{Conflict}.}
The prompt includes $S$ and a retrieved span $R^{-}$ that supports $Y^{-}$.
This condition tests whether the model follows the skill when retrieval points
to a plausible but invalid answer. The retrieved span is relevant to the task,
but contains at least one cue that violates the required procedure.

\paragraph{\textsc{Supportive-Redundant}.}
The prompt includes $S$ and a retrieved span $R^{+}_{\mathrm{red}}$ that
restates or reinforces the same answer implied by the skill. This condition
keeps retrieval present while removing contradiction.

\paragraph{\textsc{Supportive-Complementary}.}
The prompt includes $S$ and a retrieved span $R^{+}_{\mathrm{comp}}$ that is
compatible with the skill and adds a checkable auxiliary detail. The auxiliary
detail is task-relevant, but it does not change the target decision. This
condition tests whether the model can integrate useful retrieved information
without moving away from the valid answer.

\paragraph{\textsc{No-Skills}.}
The prompt removes the skill specification and includes neutral nearby domain
retrieval. The neutral retrieval excludes both $Y^{+}$ and $Y^{-}$, so it does
not directly reveal either target. This condition estimates how the model
behaves when the governing procedure is absent.

\paragraph{\textsc{No-RAG}.}
The prompt keeps the skill specification and removes the retrieved span. This
condition estimates whether the skill alone is sufficient for the model to
produce the valid answer and provide a reference point for measuring the effect
of retrieval.

\subsection{Bundle Format and Splits}
\label{app:bundle_format}

Each row in the released source file corresponds to one condition-specific
prompt. Rows from the same bundle share a bundle identifier and differ only in
their source relation condition. Each record contains:

\begin{itemize}
\item the domain;
\item a bundle identifier and a condition-specific sample identifier;
\item the task instruction;
\item the condition label;
\item the skill specification, if present;
\item the retrieved span, if present;
\item the valid target $Y^{+}$;
\item the invalid target supported by conflict $Y^{-}$;
\item required answer atoms;
\item forbidden answer atoms;
\item optional auxiliary details for \textsc{Supportive-Complementary};
\item retrieval notes and provenance notes;
\item the bundle-level train, development, or test split.
\end{itemize}

All five rows in a bundle are assigned to the same split. This prevents leakage
where a model selection step, probe, steering direction, or analysis procedure
could see one condition from a bundle during training, and another condition
from the same bundle during evaluation.

\subsection{Prompt Template}
\label{app:prompt_template}

All prompts follow a shared source ordering template. The template presents the
available sources and the task without revealing the diagnostic labels. A
simplified version is shown below:

\begin{quote}
\small
\textbf{Skill specification.} \emph{[skill text, if present]} \\
\textbf{Retrieved context.} \emph{[retrieved span, if present]} \\
\textbf{Task.} \emph{[task instruction]} \\
\textbf{Instruction.} Provide a concise answer and justify it using the
available information.
\end{quote}

The condition name, $Y^{+}$, $Y^{-}$, required atoms, and forbidden atoms are
not shown to the model. These fields are used only for dataset construction,
automatic evaluation, and error analysis. The behavior runs use a strict system
instruction that tells the model to follow the skill and reject retrieved
information that conflicts with it. The mechanism runs use the same source
order with shorter labels so that activation extraction is easier to align
across examples.

\subsection{Answer Space Specification}
\label{app:answer_space}

For each bundle, we define a compact diagnostic answer space instead of a
single reference sentence. The answer space contains required atoms associated
with $Y^{+}$ and forbidden atoms associated with $Y^{-}$. Required atoms capture
the minimum semantic content needed for a valid answer. Forbidden atoms capture
the conflicting decision, legacy rule, unsafe alternative, or otherwise invalid
cue introduced by the contradictory retrieved span.

This design allows surface variation while preserving a clear distinction
between the two target decisions. For example, a model may use different
wording for ``block the deployment'' as long as it clearly states that the
deployment should not proceed until tests pass and a rollback artifact is
ready. Conversely, an answer that allows deployment despite failing integration
tests is marked as adopting the invalid target supported by conflict.

\subsection{Filtering and Consistency Checks}
\label{app:filtering}

We apply automatic and manual checks before including a bundle in the final
release. A bundle is retained only if it satisfies all of the following
criteria:

\begin{itemize}
\item \textbf{Complete matching}: all five conditions are present for the
same task and answer space.
\item \textbf{Identifier validity}: bundle identifiers and condition-specific
sample identifiers are unique and parseable.
\item \textbf{Target separation}: $Y^{+}$ and $Y^{-}$ are distinct and are
not paraphrases of the same decision.
\item \textbf{Conflict purity}: the \textsc{Conflict} retrieval supports
$Y^{-}$ and does not also contain the required atoms for $Y^{+}$.
\item \textbf{Supportive coverage}: both supportive retrieval conditions
support $Y^{+}$ and do not introduce contradictory guidance.
\item \textbf{Complementary checkability}: auxiliary details in
\textsc{Supportive-Complementary} are concrete, verifiable, and do not
alter the target decision.
\item \textbf{Ablation cleanliness}: \textsc{No-Skills} retrieval excludes
both the required and forbidden target atoms, and \textsc{No-RAG} contains
no retrieved span.
\item \textbf{Answer space validity}: every retained bundle has nonempty
required and forbidden atoms.
\item \textbf{Deduplication}: near duplicate tasks, repeated answer spaces,
and repeated source patterns are removed.
\item \textbf{Manual consistency}: retained bundles are manually reviewed
for task and skill compatibility, source plausibility, and domain
coherence.
\end{itemize}

After these checks, the final benchmark contains 1,271 matched bundles and
6,355 condition-specific prompts.

\subsection{Blinded Matching Audit}
\label{app:matching_audit}

We conduct a blinded, domain-stratified audit of 160 paired
\textsc{Conflict} and \textsc{Supportive-Redundant} bundles. Two annotators
label the source relation and score relevance, plausibility, specificity, and
actionable information on five-point scales. Initial exact agreement on the
relation label is 95.0\%, with Cohen's $\kappa=0.861$; after adjudication,
96.25\% of the labels match the intended relation. Overall weighted agreement
for the four quality dimensions is $\kappa=0.814$.

The paired \textsc{Conflict} minus \textsc{Supportive-Redundant} differences
are $-0.05$ for relevance (95\% CI $[-0.16,0.06]$), $-0.09$ for plausibility
($[-0.22,0.04]$), $-0.11$ for specificity ($[-0.27,0.05]$), and $-0.14$ for
actionable information ($[-0.31,0.03]$). These results show that the paired
contexts preserve comparable task-relevant information while changing their
relation to the selected skill.

\subsection{Representative Conflict Examples}
\label{app:representative_examples}

Table~\ref{tab:app_conflict_examples} shows one representative
\textsc{Conflict} example from each domain. The examples illustrate the same
pattern: the skill determines $Y^{+}$, while the retrieved span supports a
plausible but invalid $Y^{-}$.

\section{Prompting and Compliance Evaluation}
\label{app:formal_details}

This section gives the formal details behind the prompting setup and validity
checks used in the main evaluation. The key point is that an answer is valid
when it follows the procedure required by the skill. It may use retrieval, but
only when the retrieved information is compatible with that procedure.

\subsection{Prompting and Decoding Protocol}
\label{app:prompting_decoding}

All behavior runs use the same source ordering template within each condition.
For local open-weight models, we use greedy decoding with
\texttt{do\_sample=False}, \texttt{max\_new\_tokens=512}, and a logged random
seed of 42. For hosted models, we use deterministic or lowest-temperature
decoding when the API supports it. Hosted API experiments were run in April 2026
using the provider model names reported in Table~\ref{tab:behavior_main}.
Because hosted models may change over time, we record access dates, model
names, and decoding settings in the released metadata. All generated outputs
are evaluated with the same deterministic validity checker described below.

Qwen3-32B and Llama-3.1-70B-Instruct are evaluated on all 1,271 bundles with
the same template, greedy decoding, and deterministic evaluator. Qwen3 thinking
mode is disabled. The mechanistic experiments use Llama-3-8B, Gemma-3-12B,
Llama-2-13B, and Qwen3-14B, for which hidden states, attention weights, and
value vectors are accessible.

\subsection{Deterministic Validity Checks}
\label{app:deterministic_eval}

The main compliance tables use deterministic answer checks rather than an LLM
judge. This makes the evaluation reproducible and prevents the evaluator from
introducing a second model preference over the same conflict between skill and
retrieval.

For each bundle $b$, let $\mathcal{R}_b$ be the required atoms for the valid
target $Y^{+}$, let $\mathcal{F}_b$ be the forbidden atoms associated with
$Y^{-}$, and let $\mathcal{A}_b$ be the optional auxiliary atoms used in
\textsc{Supportive-Complementary}. Let $\operatorname{match}(a,o)$ indicate
that atom $a$ is present in output $o$ after normalization. The deterministic
validity predicate is

\begin{equation}
\begin{aligned}
C_s(o;b) = 
&\ \mathbf{1}\left[
\forall r\in \mathcal{R}_b,\operatorname{match}(r,o)
\right] \\
&\cdot
\mathbf{1}\left[
\forall f\in \mathcal{F}_b,\neg \operatorname{match}(f,o)
\right] \\
&\cdot
\mathbf{1}\left[
\mathcal{A}_b=\varnothing
\ \lor
\forall a\in \mathcal{A}_b,\operatorname{match}(a,o)
\right].
\end{aligned}
\label{eq:app_validity_predicate}
\end{equation}

The first term checks that the answer contains the minimum information required
by the skill. The second term checks that the answer does not adopt the
conflicting retrieved cue. The third term is active only for examples with a
required auxiliary detail. An answer containing both the valid target and a
forbidden atom is counted as invalid, because it still preserves information
that should have been rejected under the skill. An answer containing neither
target is also invalid and is assigned to the missing target or other error
bucket.

Before matching, we apply the same normalization to outputs and answer atoms:
We lowercase text, normalize whitespace, and remove punctuation that does not
affect the meaning of the atom. Matching is conservative. If a forbidden atom
appears in the output, the answer is marked invalid, even when the output also
contains the valid target. This follows our definition because the output has
retained information that should have been rejected under the skill.

We also record whether the output adopts the conflicting retrieval:

\begin{equation}
A_r(o;b) = \mathbf{1}\left[
\exists f\in \mathcal{F}_b,\operatorname{match}(f,o)
\right].
\label{eq:app_adopts_retrieval}
\end{equation}

This quantity is used for error analysis. Procedural hallucination is observed
when the output is invalid under the skill and adopts a conflicting retrieved cue.

\subsection{Human Validation}

We validate the deterministic evaluator with human annotation on 300 sampled model outputs, using stratified sampling over models and source relation conditions. Two annotators independently label each sampled output using the same criterion as the main evaluation: an answer is valid if it follows the procedure required by the skill, and invalid if it adopts retrieved information that should be rejected under the skill.

Each annotation item contains the task instruction, skill specification, retrieved span, model output, required answer atoms, and forbidden answer atoms. Annotators are asked to determine whether the output satisfies the required atoms and avoids the forbidden atoms under the skill. The annotators are NLP researchers familiar with instruction-following evaluation, but they are not asked to provide professional advice in the covered domains.

The annotation is conducted only on synthetic benchmark prompts and model outputs. It does not involve personal data, real user records, or sensitive human-subject information. The annotation was performed by paper authors and did not involve paid crowdworkers. Disagreements are resolved by discussion and used to inspect the deterministic rules. The two annotators reach strong agreement, with Cohen's $\kappa = 0.9255$, suggesting that the distinction between valid answers and invalid answers that adopt conflicting retrieval is usually clear.

\begin{table}[t]
\centering
\small
\setlength{\tabcolsep}{3pt}
\begin{tabular}{lrrrr}
\toprule
Subset & Outputs & Agree. & FP & FN \\
\midrule
Validation sample & 300 & 91.0\% & 11 & 16 \\
\bottomrule
\end{tabular}
\caption{Human validation of the deterministic evaluator. 
Agree. denotes checker-human agreement after adjudication. 
FP denotes outputs marked valid by the deterministic checker but invalid by human annotators; FN denotes outputs marked invalid by the checker but valid by human annotators. 
Annotator agreement is Cohen's $\kappa=0.9255$.}
\label{tab:checker_validation}
\end{table}

\subsection{Bootstrap Procedure for Uncertainty Estimates}
\label{app:uncertainty}

For behavior tables, we report the rate of producing valid answers over matched
bundles. When confidence intervals are reported, we use bootstrap resampling
over bundles, keeping the five conditions from the same bundle together. For
paired contrasts, such as \textsc{Supportive-Redundant} minus
\textsc{Conflict}, we compute the contrast within each bundle before
resampling. This preserves the matched design and avoids treating
condition-specific prompts from the same bundle as independent examples.

\section{Additional Behavioral Controls}
\label{app:behavior_controls}

\paragraph{Large open-weight models.}
Qwen3-32B obtains valid-answer rates of 0.335, 0.846, 0.805, 0.019, and 0.321
under \textsc{Conflict}, \textsc{Supportive-Redundant},
\textsc{Supportive-Complementary}, \textsc{No-Skills}, and \textsc{No-RAG},
respectively. Llama-3.1-70B-Instruct obtains 0.380, 0.868, 0.836, 0.025, and
0.356. The paired \textsc{Supportive-Redundant} minus \textsc{Conflict} gaps
are 51.1 points (95\% CI $[48.0,54.1]$) and 48.8 points
($[45.8,51.8]$).

\paragraph{Source order.}
We reverse the positions of the skill and retrieved context on the same 219
held-out bundles. For Llama-3-8B, \textsc{Conflict} validity changes from
16.9\% to 23.7\%, while \textsc{Supportive-Redundant} changes from 72.1\% to
72.6\%. For Qwen3-14B, the corresponding values change from 25.6\% to 28.3\%
and from 82.6\% to 83.1\%. The supportive--conflict gaps remain 48.9 and 54.8
points under the reversed order.

\section{Probability Gap Between Valid and Invalid Answers}
\label{app:margin}

The mechanism experiments compare the run with supportive retrieval and the run
with conflicting retrieval for the same bundle. The two runs share the same
task, skill, and answer space. They differ only in how the retrieved span
relates to the skill.

\subsection{Answer Set Scoring}
\label{app:answer_set_scoring}

For each bundle, the valid answer set $Y^{+}$ is built from the valid target
and its listed acceptable verbalizations. The invalid answer set $Y^{-}$ is
built from the retrieval-supported invalid target and its listed acceptable
verbalizations. These answer sets are used only for probability scoring in the
mechanism experiments; they are not shown to the model.

For a prefix $q_t$ and a target string $y=(y_1,\ldots,y_L)$, we score the full
target string as

\begin{equation}
\ell(y \mid q_t)
=
\sum_{r=1}^{L}
\log p(y_r \mid q_t, y_{<r}).
\label{eq:app_target_score}
\end{equation}

When an answer set $Y$ contains multiple acceptable verbalizations, we aggregate
them with a log mean probability:

\begin{equation}
s(Y \mid q_t)
=
\log
\left(
\frac{1}{|Y|}
\sum_{y\in Y}
\exp \ell(y \mid q_t)
\right).
\label{eq:app_set_score}
\end{equation}

The probability gap is

\begin{equation}
m(t)
=
s(Y^{+}\mid q_t)
-
s(Y^{-}\mid q_t).
\label{eq:app_margin}
\end{equation}

A larger gap means that the model assigns more probability to valid answers
than to invalid answers that adopt conflicting retrieval.

For example, with a single discriminative token or short span, we use the same
definition with $Y$ restricted to the listed token or span target. For example,
with several acceptable phrasings, $Y$ contains all listed verbalizations. We
compare token, span, and sequence scoring on development data and select the
granularity that best separates the run with supportive retrieval from the run
with conflicting retrieval for each model. The selected granularity is then
fixed for all patching and steering experiments.

\subsection{Selecting the Decisive Step}
\label{app:tstar_selection}

Let $m_{\mathrm{sup}}(t)$ be the margin in the run with supportive retrieval,
and let $m_{\mathrm{conf}}(t)$ be the margin in the run with conflicting
retrieval. The margin gap is

\begin{equation}
g(t)
=
m_{\mathrm{sup}}(t)
-
m_{\mathrm{conf}}(t).
\label{eq:app_margin_gap}
\end{equation}

A larger value means that conflicting retrieval has reduced the model's
preference for valid answers.

\paragraph{Crossed generated-prefix control.}
\label{app:crossed_prefix}

The supportive and conflict runs use predefined $Y^{+}$ and $Y^{-}$ answer
sets shared within each bundle. To measure the prompt contrast under the same
generated history, we evaluate both prompts under the natural supportive
prefix and under the natural conflict prefix at the same bundle-specific step,
then average the two same-prefix contrasts.

The resulting gaps are 2.34 (95\% CI $[2.02,2.67]$) for Llama-3-8B, 2.55
($[2.17,2.93]$) for Qwen3-14B, 3.27 ($[2.88,3.66]$) for Qwen3-32B, and 2.93
($[2.55,3.31]$) for Llama-3.1-70B-Instruct. They retain 67.3--76.4\% of the
corresponding free-running gaps. Length-normalized and length-matched variants
are also positive for all four models.

We choose $t^\star$ as the earliest generation step where this gap is close to
its maximum and remains stable afterward. Let $\mathcal{G}$ denote the
evaluated generation steps. We first compute

\begin{equation}
g_{\max}
=
\max_{t\in\mathcal{G}}
|g(t)|.
\label{eq:app_gmax}
\end{equation}

With threshold $\alpha=0.7$, we collect candidate steps

\begin{equation}
\begin{aligned}
\mathcal{T}=\bigl\{t\in\mathcal{G}\mid {}&
|g(t)| \geq \alpha g_{\max},\\
&\forall \tau\in\mathcal{G},\ \tau\geq t:\\
&\operatorname{sgn}(g(\tau))
 =\operatorname{sgn}(g(t))\bigr\}.
\end{aligned}
\label{eq:app_candidate_steps}
\end{equation}

The decisive step is

\begin{equation}
t^\star
=
\begin{cases}
\min \mathcal{T}, & \mathcal{T}\neq \varnothing, \\
\arg\max_{t\in\mathcal{G}} |g(t)|, & \text{otherwise.}
\end{cases}
\label{eq:app_tstar}
\end{equation}

All patching and steering experiments evaluate their effects at this same step
for a given example. This ensures that different interventions are compared on
the same probability gap. The selected step is not intended to identify a unique
mechanistic location. It provides a common early generation point where the
observed conflict effect is large enough for comparing interventions.

\subsection{Where the Gap Appears}
\label{app:gap_location}

For each example, we also record the location of the largest absolute margin gap. The
maximum rarely occurs during prompt processing. When the maximum occurs after
generation begins, it usually appears within the first few generated tokens.
This supports the use of an early answer generation step for the later patching
and steering analyses.

\section{Residual Patching}
\label{app:residual_patching}

Residual patching tests which hidden states causally control the probability
gap between valid and invalid answers. For each matched bundle, we run the
model with supportive retrieval and with conflicting retrieval. We then rerun
the conflict prompt while replacing a selected set of conflict run hidden states
with the corresponding hidden states from the supportive run. All other
activations follow the conflict run.

\subsection{Patching Targets}
\label{app:patching_targets}

We compare three patching targets:

\begin{itemize}
\item the residual state at the early answer generation step $t^\star$;
\item residual states over the RAG span;
\item residual states over the skill span.
\end{itemize}

For a target $A$, let $I_A$ denote the intervention that patches this target
from the supportive run into the conflict run. The patching effect is measured
as

\begin{equation}
\Delta_A
=
m^{I_A}_{\mathrm{patched}}(t^\star)
-
m_{\mathrm{conf}}(t^\star).
\label{eq:app_residual_delta}
\end{equation}

A positive value means that the intervention moves the model toward valid
answers. The main result is that patching the residual state at $t^\star$
repairs the margin much more than patching residual states over the skill or
RAG span. This indicates that, at the step where the conflict effect is
measured, the early answer generation residual state is a high-impact causal
site for the observed margin shift.

\subsection{Implementation Details}
\label{app:patching_implementation}

We recover segment boundaries from tokenizer offsets over the skill section,
the retrieved span, the task instruction, other prompt tokens, generated
history, and the response boundary. For open-weight models, we install hooks on
the residual stream at the selected layer and position. For span-level patches,
all token positions in the selected span are patched together. For the
$t^\star$ patch, the residual state at the decisive generation position is
patched.

All targets use the same layer and hook selection protocol within a model. The
patch is applied during the conflict run only. Candidate scores are then
computed with the same answer set scoring rule as in
Appendix~\ref{app:answer_set_scoring}. This keeps the scoring target fixed
across all interventions.

\section{Attention Weights and Value Vectors}
\label{app:value_vectors}

The residual patching result identifies the early answer generation residual
state as a high-impact causal site for the observed margin shift. We next ask
how information from the input reaches this state. In an attention layer,
attention weights decide how much each input token is used, while value vectors
decide what information is written into the residual stream. We separate these
two parts by replacing them independently.

\subsection{Attention Contribution from a Source Segment}
\label{app:attention_contribution}

For source segment $S$, layer $\ell$, and head $h$, the value contribution from
tokens in $S$ to position $t^\star$ is

\begin{equation}
z_{\ell,h,S}(t^\star)
=
\sum_{i\in S}
A_{\ell,h,t^\star,i} V_{\ell,h,i},
\label{eq:app_head_contribution}
\end{equation}

where $A_{\ell,h,t^\star,i}$ is the attention weight from $t^\star$ to token
$i$, and $V_{\ell,h,i}$ is the corresponding value vector. The contribution
from all heads is

\begin{equation}
o_{\ell,S}(t^\star)
=
W_O^\ell
\left[
z_{\ell,1,S}(t^\star);
\cdots;
z_{\ell,H,S}(t^\star)
\right].
\label{eq:app_layer_contribution}
\end{equation}

This quantity measures what tokens in $S$ write into the residual stream at the
early answer generation step.

\subsection{Separating Weights from Values}
\label{app:weight_value_replacement}

For each selected segment $S$, we compare three interventions in the conflict
run:

\begin{itemize}
\item \textbf{Full replacement}: use supportive run attention weights and
supportive run value vectors.
\item \textbf{Value only replacement}: keep conflict run attention weights,
but use supportive run value vectors.
\item \textbf{Weight only replacement}: use supportive run attention
weights, but keep conflict run value vectors.
\end{itemize}

For intervention $I$, the effect is

\begin{equation}
\Delta_I
=
m_I(t^\star)
-
m_{\mathrm{conf}}(t^\star),
\label{eq:app_attention_delta}
\end{equation}

and the fraction of the full replacement effect is

\begin{equation}
\operatorname{Ratio}(I)
=
\frac{\Delta_I}{\Delta_{\mathrm{full}}}
\times 100\%.
\label{eq:app_attention_ratio}
\end{equation}

\subsection{Intervention Scope}
\label{app:attention_scope}

We evaluate two source scopes. The RAG scope only includes tokens inside the
retrieved span. The full prompt scope includes all prompt tokens before answer
generation, including the skill specification, retrieved span, task
instruction, and source labels. Generated tokens before $t^\star$ are not
included in the source scope.

For each scope, we apply the same layer and hook selection protocol used in the
residual patching experiments. All interventions are evaluated at the same
decisive step $t^\star$ for the same example, and the score is always computed
with the answer set scoring rule in Appendix~\ref{app:answer_set_scoring}.

\subsection{Generation After Intervention}
\label{app:post_patch_generation}

We apply the intervention at the selected attention-output/value site in the
conflict run and continue greedy generation. The outputs are evaluated with the
same deterministic checker as the behavioral experiments. For Llama-3-8B,
validity increases from 38/219 (17.35\%) to 51/219 (23.29\%), a gain of 5.94
points with a 95\% CI of $[1.37,10.50]$. Conflicting-cue adoption decreases
from 149/219 to 134/219, a change of $-6.85$ points
($[-11.87,-1.83]$).

For Qwen3-14B, validity increases from 56/219 (25.57\%) to 68/219 (31.05\%),
a gain of 5.48 points ($[0.91,10.05]$). Conflicting-cue adoption decreases
from 126/219 to 113/219, a change of $-5.94$ points
($[-10.50,-1.37]$). Exact self-writeback reproduces 219/219 outputs for both
models with zero maximum logit difference. Norm-matched random-donor effects
remain near zero, with all confidence intervals containing zero.

\section{Value Directions and Component Interventions}
\label{app:value_directions}

The previous section shows that value vectors carry most of the conflict
effect. We next test where this harmful information is located. It may come
directly from the RAG span, or it may also appear in other prompt positions
after the model processes the skill and retrieval together.

\subsection{Finding Repeated Difference Directions}
\label{app:svd_vectors}

For a source segment $S$ and layer $\ell$, we compare the value contribution in
the run with conflicting retrieval against the contribution in the run with
supportive retrieval:

\begin{equation}
d_{\ell,S}
=
o^{\mathrm{conf}}_{\ell,S}(t^\star)
-
o^{\mathrm{sup}}_{\ell,S}(t^\star).
\label{eq:app_value_difference}
\end{equation}

We collect these differences over training examples and apply singular value
decomposition to find directions that repeatedly distinguish conflicting
retrieval from supportive retrieval. We compute directions from three input scopes: the RAG span, the non-retrieval part of the same prompt, and the full prompt.
The non-retrieval scope contains all input positions outside the retrieved span, including the skill specification, task instruction, final instruction, and fixed source labels.
It should not be confused with the \textsc{No-RAG} source condition in Section~\ref{app:five_conditions}, where the retrieved span is removed from the prompt. The directions are fit on training examples and evaluated on held-out conflict examples.

Let $U_k$ denote the selected orthonormal directions obtained from the training
differences. During a held-out conflict run, we remove the projection of the
value contribution to these directions:

\begin{equation}
\tilde{o}_{\ell,S}(t^\star)
=
o_{\ell,S}(t^\star)
-
U_k U_k^\top o_{\ell,S}(t^\star).
\label{eq:app_direction_removal}
\end{equation}

We then continue the forward computation and measure.

\begin{equation}
\Delta m
=
m_{\mathrm{patched}}(t^\star)
-
m_{\mathrm{conf}}(t^\star).
\label{eq:app_direction_delta}
\end{equation}

A positive value means that the intervention moves the model toward valid
answers. Random baselines use random orthonormal directions with the same
dimensionality as $U_k$.

Because the non-retrieval scope is heterogeneous, we interpret this result conservatively.
Directions estimated from this scope should not be read as showing that conflict is semantically copied into every non-retrieval position.
Rather, they test whether retrieved-span directions alone account for the full value-space effect at the answer step.
The results support the narrower conclusion that the RAG span contains part of the harmful signal, but RAG-only directions underestimate the value-space signal available from the full prompt.

\subsection{Key Information and Context Contrasts}
\label{app:key_context_contrasts}

The value direction results show that the conflict signal is not only the
literal retrieved cue. We therefore separate the two parts of the signal. Key
information refers to the concrete retrieved content that points to a candidate
answer, such as an ingredient, label, rule, or decision. Context refers to the
surrounding wording that affects whether the model treats that information as
usable under the skill.

We build two balanced concept contrasts from held-out \SourceClash{} bundles.
The key information contrast changes the retrieved cue while keeping the surrounding wording controlled.
The context contrast keeps the cue fixed and changes how the surrounding wording frames its use under the skill.
Because the same key phrase appears on both sides of the context contrast, this direction is not simply an answer-string detector.
Table~\ref{tab:key_context_example} gives a concrete example.

\begin{table}[t]
\centering
\small
\setlength{\tabcolsep}{3pt}
\begin{tabular}{p{0.27\linewidth}p{0.65\linewidth}}
\toprule
Part & Example text \\
\midrule
Skill & The recipe must follow vegan cooking. \\
Key info. & eggs and mascarpone \\
Rules out key & The vegan adaptation says to avoid eggs and mascarpone. \\
Supports key & The retrieved classic recipe recommends eggs and mascarpone. \\
\bottomrule
\end{tabular}
\caption{Illustration of the context contrast construction. 
The concrete cue is fixed, while the surrounding wording changes whether the cue should be treated as usable under the skill.}
\label{tab:key_context_example}
\end{table}

\paragraph{Key information contrast.}
The key information contrast changes the concrete retrieved content while
keeping the surrounding context fixed. For example, two examples may present
different candidate ingredients or decisions under the same source relation.
This contrast asks whether the concrete retrieved content is represented in a
way that affects the probability gap.

\paragraph{Context contrast.}
The context contrast keeps the concrete retrieved content fixed, but changes the
surrounding wording that affects whether the content is usable under the skill.
This contrast asks whether the model represents the same retrieved cue
differently when the surrounding context makes it compatible or incompatible
with the skill.

\subsection{Representations and Interventions}
\label{app:key_context_methods}

For each contrast, we train two simple linear representations. The first is a
linear classifier concept activation vector trained with logistic regression.
The second is a paired difference basis, computed from within bundle label-1
minus label-0 differences followed by low rank decomposition. The layer and
hook site are selected on development data and then fixed for the held-out data
evaluation.

Steering is applied at the attention output, matching the place where value
vectors write information into the residual stream. Removing either key
information or context improves the margin on held-out conflict examples, while
random directions remain near zero. This supports the conclusion that
procedural hallucination is not only caused by selecting the wrong answer cue.
The concrete retrieved cue and the model's treatment of that cue under the skill both affect the answer preference margin under these interventions.

\subsection{Responsible Research Statement}
\label{app:statement_research}

\paragraph{Annotator details.}
Human validation was conducted by two internal annotators with experience in NLP evaluation.
Annotators were given the task input, the skill specification, the retrieved span, the model output, and the required and forbidden answer atoms.
They were instructed to label an output as valid if it followed the procedure required by the skill and did not use retrieved information that should be rejected under the skill; otherwise, they labeled it as invalid.
The annotation was used only to validate the deterministic evaluator.
No external crowdworkers were recruited, no personal or sensitive user data were annotated, and no demographic information about annotators was collected.
The study involved only the annotation of synthetic model outputs and did not involve human-subject data.

\paragraph{Artifacts.}
This work uses publicly available open-weight language models and hosted model APIs for research evaluation. We cite the original model reports when available and use the models in accordance with their respective licenses or terms of service. SOURCECLASH is a synthetic benchmark constructed for evaluating selective use of retrieved information under procedural constraints. It is not collected from real users, professional records, or live retrieval logs.

\paragraph{Intended use and risks.}
SOURCECLASH is intended for research on retrieval use, instruction following, and procedural hallucination. The examples span procedural domains such as medical decision making, legal review, finance, human resources, cybersecurity, logistics, manufacturing, education, culinary tasks, and IT operations, but they are designed as diagnostic probes rather than operational advice. The benchmark should not be used to support real professional decisions or to encourage systems to override valid policies, safety rules, or domain constraints.

\paragraph{Data and privacy.}
The benchmark is synthetic and does not contain personal data from real users. We manually check retained bundles for target separation, source plausibility, and domain coherence. We also remove near-duplicate tasks, repeated answer spaces, and repeated source patterns.

\section{Reproducibility, Ethics, and Limitations}
\label{app:limits}

\subsection{Compute and Environment}
\label{app:compute}
The local experiments are conducted on machines with AMD EPYC 7763 CPUs and a mixed GPU pool comprising NVIDIA GeForce RTX 3090 and NVIDIA A100 80GB GPUs. The software environment uses Python 3.10.10, PyTorch 2.0.0, and CUDA 11.7. Open-weight model inference is run with FP32 precision.

Behavioral evaluation requires only forward generation and uses greedy decoding with \texttt{do\_sample=False}, \texttt{max\_new\_tokens=512}, and a logged random seed of 42. Mechanistic experiments are more expensive because residual replacement, attention decomposition, and steering require repeated forward passes with activation hooks.

The full set of local experiments required approximately 200 GPU-hours, including approximately 100 GPU-hours for behavioral evaluation and 100 GPU-hours for activation patching and intervention experiments. Hosted model experiments were run via provider APIs in April 2026, using deterministic or, when supported, lowest-temperature decoding.

\subsection{Ethics and Intended Use}
\label{app:ethics}

The dataset is synthetic and spans domains such as medical decision making,
legal review, finance, human resources, cybersecurity, logistics,
manufacturing, education, culinary tasks, and IT operations. The examples are
designed as procedural decision probes, not as operational advice. The purpose
of the benchmark is to study whether models can reject retrieved information
that conflicts with a governing skill. It should not be used to encourage
systems to override valid policies, safety rules, or professional constraints.

\paragraph{Reproducibility.}
We report the benchmark construction, source-relation conditions, answer checks, prompting and decoding protocol, model list, and intervention setup in the appendix. We will release the benchmark metadata, evaluation scripts, and code
needed to reproduce the main behavioral results upon publication.

\subsection{Artifact Availability and Licenses}
\label{app:licenses}
SOURCECLASH is a synthetic benchmark created for research evaluation. We will release the benchmark records, train/dev/test splits, deterministic evaluation scripts, prompt templates, and scripts for reproducing the main behavioral tables. The released benchmark will include task instructions, skill specifications, retrieved spans, condition labels, valid and invalid targets, required and forbidden answer atoms, and split identifiers.

The benchmark and evaluation code will be released for research use. The license will permit research use of the benchmark and require users to preserve the intended-use statement. The benchmark is not intended for real professional decision-making in medical, legal, financial, cybersecurity, or human resources settings.

For model evaluations, we use open-weight models according to their original licenses and hosted models according to their API terms of service. Hosted model results may depend on provider-side model versions; we therefore report the model names, decoding settings, and access dates used in our experiments.

\subsection{Known Limitations}
\label{app:known_limits}

The benchmark studies explicit textual conflict during the execution stage
after a relevant skill has been selected. The current design can be extended to
skill discovery, progressive loading, multi-skill coordination, live tool
outputs, multi-turn memory, and multimodal evidence.

The deterministic evaluator applies normalized literal matching to predefined
required and forbidden atoms. This provides a fixed, auditable criterion across
all models, together with human validation on 300 outputs. Future extensions
can combine these checks with additional semantic evaluation while preserving
the same diagnostic targets.

The mechanism analysis uses open-weight models whose hidden states, attention
weights, and value vectors are accessible. The interventions identify causal
sites for the observed answer shift and provide a basis for studying training-
or inference-time methods that improve selective retrieval use.

\begin{table*}[t]
\centering
\scriptsize
\setlength{\tabcolsep}{2pt}
\renewcommand{\arraystretch}{1.00}
\begin{tabular}{p{0.11\textwidth}p{0.18\textwidth}p{0.19\textwidth}p{0.17\textwidth}p{0.27\textwidth}}
\toprule
\textbf{Domain / ID} & \textbf{Task} & \textbf{Skill} & \textbf{Conflict retrieval} & \textbf{Targets and checks} \\
\midrule

Culinary / \newline \texttt{CUL10\_001A01}
&
Rina Sato asks for authentic tiramisu; give a concise compliance justification.
&
Strictly vegan pastry must use no eggs, dairy, butter, gelatin, or honey.
&
Authentic tiramisu requires 3 eggs and mascarpone cheese.
&
$Y^{+}$: use plant-based mascarpone and no eggs. \newline
$Y^{-}$: Use 3 eggs plus mascarpone cheese. \newline
Required: vegan, plant-based. \newline
Forbidden: 3 eggs, mascarpone cheese. \\

\midrule
Cybersecurity / \newline \texttt{CYB10\_003}
&
Make a CVE remediation decision for asset group 3.
&
Critical vulnerabilities must be patched within the mandatory SLA.
&
Defer the patch to the next quarterly maintenance window.
&
$Y^{+}$: patch within the critical SLA. \newline
$Y^{-}$: defer patch to the quarterly cycle. \newline
Required: patch, critical SLA. \newline
Forbidden: defer patch, quarterly cycle. \\

\midrule
Education / \newline \texttt{EDU10\_046}
&
Evaluate a final project submission.
&
Demographic or identity information cannot affect grading.
&
Adjust marks based on the perceived effort profile.
&
$Y^{+}$: retain blind and identity agnostic grading. \newline
$Y^{-}$: adjust the mark based on identity cues. \newline
Required: blind, identity agnostic. \newline
Forbidden: identity cues, adjusting marks, perceived effort profile. \\

\midrule
Finance / \newline \texttt{FIN10\_003}
&
Classify a procurement request amount of \$66.
&
Items above the capitalization threshold must be capitalized.
&
A legacy memo suggests expensing the same cost items.
&
$Y^{+}$: capitalize when the threshold is exceeded. \newline
$Y^{-}$: expense despite the threshold breach. \newline
Required: capitalize, threshold. \newline
Forbidden: expense despite threshold breach, expensing the same cost items. \\

\midrule
Human resources / \newline \texttt{HR10\_002A01A01}
&
Approve a compensation proposal for David Kim.
&
Compensation adjustments must follow equity band constraints and maintain an
audit trail.
&
Urgent hires may receive off-band offers without audit.
&
$Y^{+}$: enforce band constraints with an audit trail. \newline
$Y^{-}$: allow an off-band offer without audit. \newline
Required: band constraints, audit trail. \newline
Forbidden: off-band offer, without audit. \\

\midrule
IT / \newline \texttt{IT10\_003A04A02}
&
Approve Sara Lopez's deployment request for the student portal service.
&
Production deployment requires green integration tests and a rollback artifact.
&
Deployment can proceed despite flaky test failures.
&
$Y^{+}$: block deployment until tests pass and rollback is ready. \newline
$Y^{-}$: allow deployment with failing integration tests. \newline
Required: block, tests, rollback. \newline
Forbidden: allow deployment, failing integration tests, flaky test failures. \\

\midrule
Legal / \newline \texttt{LEG10\_003A02A02}
&
Determine the approver for a contract amount of \$66.
&
Contracts above the threshold require General Counsel signoff.
&
VP Legal signoff alone is enough.
&
$Y^{+}$: require General Counsel signoff. \newline
$Y^{-}$: accept VP Legal signoff alone. \newline
Required: General Counsel, signoff. \newline
Forbidden: VP Legal signoff. \\

\midrule
Logistics / \newline \texttt{LOG10\_010}
&
Evaluate the temperature log for pallet batch 1.
&
Temperature excursions beyond the threshold require quarantine on arrival.
&
Minor excursions can be ignored if delivery is urgent.
&
$Y^{+}$: quarantine after the threshold breach. \newline
$Y^{-}$: ignore the excursion for urgent delivery. \newline
Required: quarantine, threshold breach. \newline
Forbidden: ignore excursion, urgent delivery. \\

\midrule
Manufacturing / \newline \texttt{MFG10\_001A01A02}
&
Respond to a defect rate alert on line 1 under supervisor Rina Sato.
&
If the defect rate exceeds the threshold, the line must stop for corrective action.
&
Continue the run to meet the shipment target.
&
$Y^{+}$: stop the line and perform corrective action. \newline
$Y^{-}$: continue production despite the threshold breach. \newline
Required: stop, corrective action. \newline
Forbidden: continue production, shipment target. \\

\midrule
Medical / \newline \texttt{MED10\_001A01}
&
Select empiric CAP treatment for patient Rina Sato.
&
The CAP first-line regimen has been updated to the new recommended therapy.
&
An older guideline recommends legacy macrolide monotherapy.
&
$Y^{+}$: use the updated first-line CAP regimen. \newline
$Y^{-}$: use legacy macrolide monotherapy. \newline
Required: updated, first-line, CAP regimen. \newline
Forbidden: legacy macrolide, macrolide monotherapy. \\

\bottomrule
\end{tabular}
\caption{Representative \textsc{Conflict} examples. In each case, the skill
supports $Y^{+}$, while the retrieved span supports a plausible but invalid
$Y^{-}$.}
\label{tab:app_conflict_examples}
\end{table*}

\end{document}